\def\year{2022}\relax
\documentclass[letterpaper]{article} 
\usepackage{aaai22}  
\usepackage{times}  
\usepackage{helvet}  
\usepackage{courier}  
\usepackage[hyphens]{url}  
\usepackage{graphicx} 
\usepackage{natbib}  
\usepackage{caption} 
\DeclareCaptionStyle{ruled}{labelfont=normalfont,labelsep=colon,strut=off} 
\usepackage{algorithm}
\usepackage{algorithmic}

\usepackage{amsfonts}       
\usepackage{lipsum}
\usepackage{amsmath}
\usepackage{multirow}
\usepackage{subfigure}
\usepackage{color}
\usepackage{CJKutf8}
\usepackage{enumitem}
\usepackage{textcomp}
\usepackage{microtype}
\usepackage{subfigure}
\usepackage{booktabs} 

\usepackage{newfloat}
\usepackage{listings}
\floatstyle{ruled}
\newfloat{listing}{tb}{lst}{}
\floatname{listing}{Listing}
\title{DeepThermal: Combustion Optimization for \\Thermal Power Generating Units Using Offline Reinforcement Learning}
\author{
    Xianyuan Zhan\textsuperscript{\rm 1}\thanks{Equal contribution. Xianyuan Zhan is the corresponding author. This research was done at JD Technology.}, Haoran Xu\textsuperscript{\rm 2,3,4 $*$}, Yue Zhang\textsuperscript{\rm 2,3}, Xiangyu Zhu\textsuperscript{\rm 2,3}, Honglei Yin\textsuperscript{\rm 2,3}, Yu Zheng\textsuperscript{\rm 2,3,4}
}
\affiliations{
    \textsuperscript{\rm 1}
    Institute for AI Industry Research (AIR), Tsinghua University, Beijing, China\\

	\textsuperscript{\rm 2}
	JD iCity, JD Technology, Beijing, China\\
	
	\textsuperscript{\rm 3}
	JD Intelligent Cities Research, Beijing, China\\
	
	\textsuperscript{\rm 4}
	Xidian University, Xi'an, China\\ 

    \{zhanxianyuan, ryanxhr, zhangyuezjx, zackxiangyu, yinhonglei93\}@gmail.com, msyuzheng@outlook.com\\
}

\usepackage{bibentry}

\begin{document}

\maketitle

\begin{abstract}
	Optimizing the combustion efficiency of a thermal power generating unit (TPGU) is a highly challenging and critical task in the energy industry. 
	We develop a new data-driven AI system, namely DeepThermal, to optimize the combustion control strategy for TPGUs. At its core, is a new model-based offline reinforcement learning (RL) framework, called MORE, which leverages historical operational data of a TGPU to solve a highly complex constrained Markov decision process problem via purely offline training. 
	In DeepThermal, we first learn a data-driven combustion process simulator from the offline dataset. The RL agent of MORE is then trained by combining real historical data as well as carefully filtered and processed simulation data through a novel restrictive exploration scheme.	
	DeepThermal has been \textbf{successfully deployed} in four large coal-fired thermal power plants in China. Real-world experiments show that DeepThermal effectively improves the combustion efficiency of TPGUs.
	We also report the superior performance of MORE by comparing with the state-of-the-art algorithms on the standard offline RL benchmarks.
\end{abstract}

\section{Introduction}

Thermal power generation forms the backbone of the world's electricity supply and plays a dominant role in the energy structure of many countries. For example, there are more than 2,000 coal-fired thermal power plants in China, contributing to more than 60\% of all electricity generated in the country.
Every year, thermal power plants across the world consume an enormous amount of non-renewable coal and cause serious air pollution issues. 
How to improve the combustion efficiency of a \textit{thermal power generating unit} (TPGU) has been a critical problem for the energy industry for decades. 
Solving this problem has huge economic and environmental impacts. 
For instance, by only improving \textbf{0.5\%} of combustion efficiency of a 600 megawatt (MW) TPGU, a power plant can save more than 4000 tons of coal and reduce hundreds of tons of emissions (e.g. carbon dioxides $\mathrm{CO_2}$ and nitrogen oxides $\mathrm{NO_x}$) a year.

After decades of development and technology advances, most modern TPGUs can achieve a combustion efficiency ranging from 90\% to 94\%. Further improving the combustion efficiency of TPGUs, especially through system control aspect is becoming an extremely challenging task. The difficulties arise from several aspects. First, TPGUs are highly complex and large systems, which contain lots of equipment, huge amounts of sensors and complicated operation mechanisms. 
The involvement of the large number of safety constraints and domain knowledge further exacerbates the difficulty of the task. 
Lastly, it is desirable to achieve long-term optimization with multiple objectives, such as increasing combustion efficiency while reducing $\mathrm{NO_x}$ emission. 
All these factors and requirements result in an extremely difficult problem that has not been well solved after decades of effort.
Currently, most coal-fired thermal power plants still use semi-automatic control systems, and their control heavily depends on the experience and expertise of human operators. 

Conventional industrial control optimization approaches, such as the widely used PID controller \cite{astrom2006advanced} and model predictive control (MPC) algorithms \cite{garcia1989model}, neither have sufficient expressive power nor scale with the increase of problem size. When facing large and complex systems, these methods will have unavoidable modeling complexity and time cost to obtain the optimal solutions.
Hence most existing combustion optimization approaches decompose the TPGU into individual small sub-systems that only optimize a limited amount of state and control variables \cite{kalogirou2003artificial,lee2007neural,ma2011neural,liu2014integrating}. 
The recent advances of deep reinforcement learning (RL) provide another promising direction. Deep RL leverages expressive function approximators and has achieved great success in solving complex tasks such as games \cite{mnih2015human,silver2017mastering} and robotic control \cite{levine2016end}. However, all these achievements are restricted to the online setting, where agents are allowed to have unrestricted interaction with real systems or perfect simulation environments. 	
In real-world industrial control scenarios, an algorithm may never get the chance to interact with the system at the training stage. A problematic control policy can lead to disastrous consequences to system operation. 
Besides, most real-world industrial systems are overly complex or partially monitored by sensors, which makes it impossible to build a high-fidelity simulator.

Fortunately, industrial systems like TPGUs have long-term storage of the operational data collected from sensors,
and the recently emerged offline RL provides an ideal framework for our problem. 
Offline RL focuses on training RL policies from offline, static datasets without environment interaction. The main difficulty of offline RL tasks is the \textit{distributional shift} issue \cite{kumar2019stabilizing}, which occurs when the learned policies make counterfactual queries on unknown out-of-distribution (OOD) data samples, causing non-rectifiable exploitation error during training.
The key insight of recent offline RL algorithms \cite{fujimoto2019off,kumar2019stabilizing,wu2019behavior,yu2020mopo} is to restrict policy learning stay close to the data distribution. However, these methods are over-conservative, hindering the chance to surpass a sub-optimal behavior policy.

In this work, we develop a new data-driven AI system, namely \textbf{DeepThermal}
(Chinese name:\begin{CJK*}{UTF8}{gbsn}
	深燧
\end{CJK*}), 
to optimize the combustion efficiency of real-world TPGUs. DeepThermal constructs a data-driven combustion process simulator to facilitate RL training.
The core of DeepThermal is a new model-based offline RL framework, called \textbf{MORE}, which is capable of leveraging both logged datasets and an imperfect simulator to learn a policy under safety constraints and greatly surpass the behavior policy. DeepThermal has already been successfully deployed in four large coal-fired thermal power plants in China. 
Real-world experiments show that the optimized control strategies provided by DeepThermal effectively improve the combustion efficiency of TPGUs. Extensive comparative experiments on standard offline RL benchmarks also demonstrate the superior performance of MORE against the state-of-the-art offline RL algorithms.

\section{Overview}

\subsection{Operation Mechanisms of TPGUs}
\label{sec:mechanism}
\begin{figure}[t]
	\includegraphics[width=0.46\textwidth]{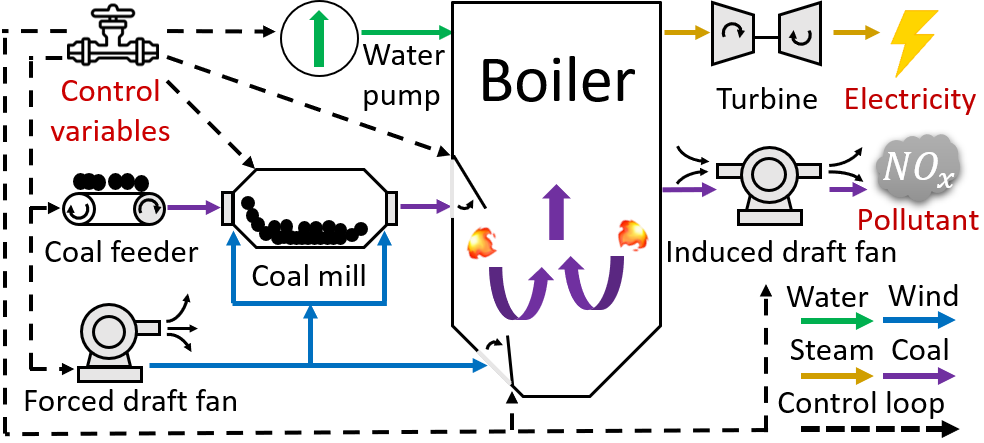}
	\caption{Illustration of operation mechanisms of a TPGU} 
	\label{fig.boiler}
\end{figure}

Thermal power generating unit converts the chemical energy of the coal to electric power. The power generation process of a TPGU is highly complicated involving three major stages (see Figure \ref{fig.boiler}). 1) \textbf{Coal pulverizing stage}: Coals from the coal-feeders are pulverized to fine-grained particles by coal mills before outputting to the burner. To ensure complete combustion, many control operations need to be properly performed, e.g. amount of coal should meet the demand load; valves of the cold and hot air blowers (primary blowers) are adjusted to ensure suitable primary air temperature. 2) \textbf{Burning stage}: Pulverized coals and air from the secondary blower are injected through 20$\sim$48 locations of the burner (depending on the specific structure of the burner). The valves of the secondary blower at each injection location need to be precisely controlled to allow a large fireball to form at the center of the burner, facilitating complete combustion. Safety and regulatory issues need also be guaranteed, such as maintaining negative internal pressure and pollutants generated below a certain level. 3) \textbf{Steam circulation stage}. The burner vaporizes water in the boiler and generates high-temperature, high-pressure steam, which drives a steam turbine to generate electricity satisfying demand load. The steam generated needs to satisfy multiple temperature and pressure requirements, which are controlled by the valves of the induced draft fan, and the amount of cooling water used, etc.

Optimizing the combustion efficiency of a TPGU involves 70$\sim$100 major continuous control variables and the chemical properties of the coal, which is extremely challenging.

\subsection{Preliminaries}

We model the combustion optimization problem for TPGUs as a Constrained Markov Decision Process (CMDP) \cite{altman1999constrained}, which augments the standard MDP with multiple safety constraints.
A CMDP is represented by a tuple $(\mathcal{S}, \mathcal{A}, T, r, c_{1:m}, \gamma)$, where $\mathcal{S}$ and $\mathcal{A}$ denote the state and action spaces, $T(s_{t+1}|s_t, a_t)$ denotes the transition dynamics. $r(s_t,a_t)>0$ is the reward function and $c_{1:m}(s_t,a_t)$ are $m$ cost functions. $\gamma\in(0,1)$ is the discount factor. A policy $\pi(s)$ is a mapping from states to actions.
In our problem,
the state, action, reward and costs are set as follows. \\
\textbf{States $\mathcal{S}$}: We use the chemical property of the coal and sensor data that relevant to the combustion process of a TPGU as states, including 
temperature, pressure, wind, and water volume as well as other sensor readings of different stages in the combustion process described in previous section. \\
\textbf{Actions $\mathcal{A}$}: We consider all the key control variables that impact combustion process in a TPGU as actions, such as the adjustment of valves and baffles. All actions are continuous. \\
\textbf{Reward function $r$}: We model the reward as a weighted combination of combustion efficiency $\mathit{Effi}$ and reduction in $\mathrm{NO_x}$ emission $\mathit{Emi}$, i.e. $r_t = \alpha_r \mathit{Effi}_t + (1-\alpha_r)\mathit{Emi}_t$.
We set $\alpha_r=0.8$ in our deployed systems, as improving combustion efficiency is the primary concern for many power plant. \\
\textbf{Cost functions $c_{1:m}$}: We model a series of safety constraints as costs, such as load, internal pressure, and temperature satisfaction. Violating these constraints will lead to a positive penalty value. We denote a weighted combination of costs as $\tilde{c}(s,a)=\sum_{i=1}^{m}\alpha_c^i c_i(s,a)$, where $\alpha_c^{1:m}$ are set according to experts' opinion.

In our 
problem, we assume no interaction with the actual TPGU and only have a static historical operational dataset $\mathcal{B}$ = ${(s,a,s',r,c_{1:m})}$, generated by unknown behavior policies from TPGU operators.
Our goal is to learn a policy $\pi^{*}(s)$ from $\mathcal{B}$ that maximizes the expected discounted reward $R(\pi) = \mathbb{E}_{\pi}[\sum_{t=0}^{\infty} \gamma^{t} r(s_t, a_t)]$ while controlling 
the expected discounted combined costs $C(\pi) = \mathbb{E}_{\pi}[\sum_{t=0}^{\infty} \gamma^{t} \tilde{c}(s_t, a_t)]$ below a threshold $l$, mathematically:
\begin{equation} 
\label{eq:1}
 \pi^{*} = \arg \max_{\pi}\ R(\pi) \quad
 \mathrm{s.t.}\quad C(\pi) \leq l
\end{equation}


\begin{figure}[t]
	\includegraphics[width=0.325\textwidth]{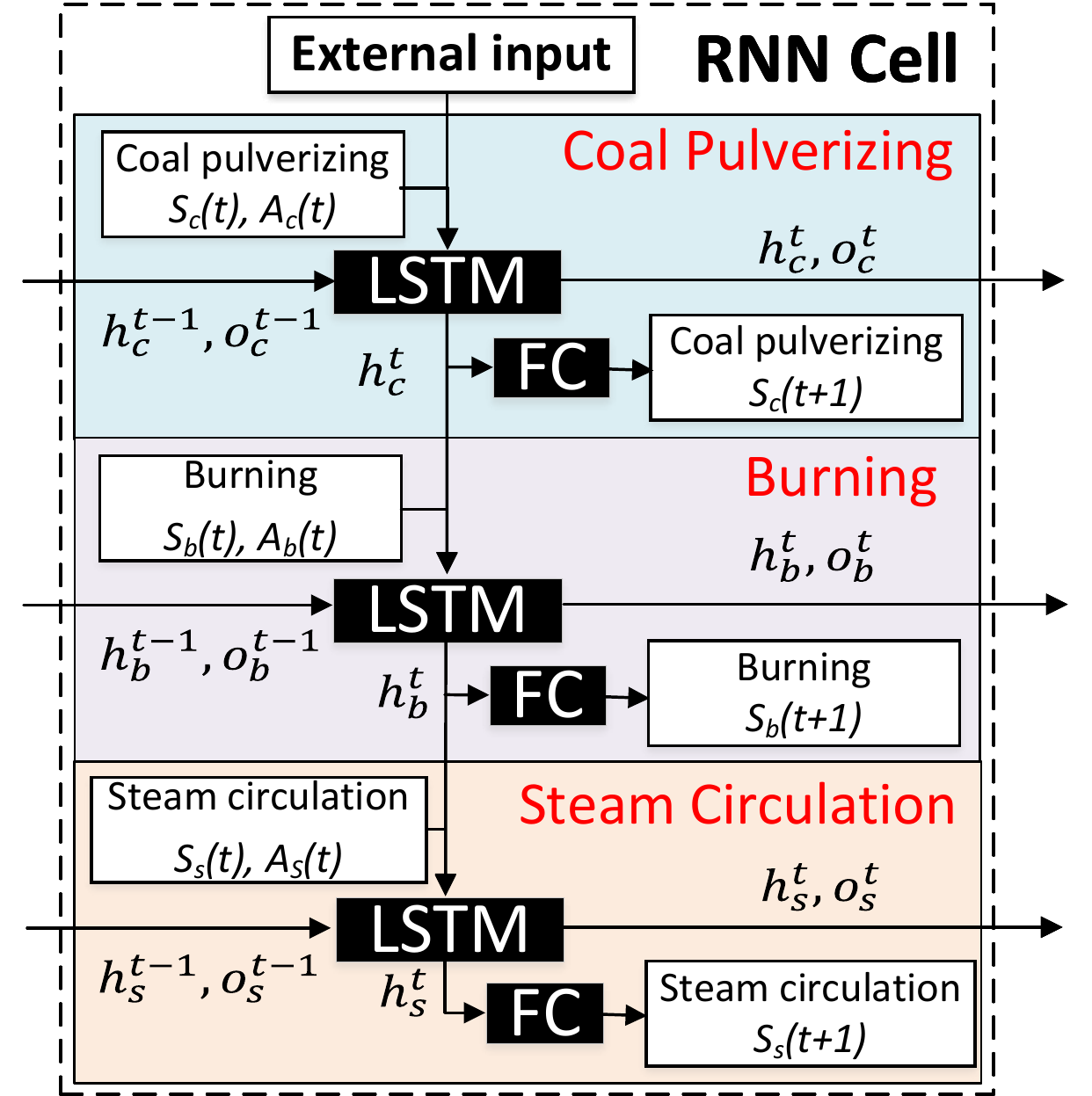}
	\centering
	\caption{Design of the combustion process simulator}
	\label{fig.simulator}
\end{figure}

\section{Combustion Process Simulator}

DeepThermal learns a data-driven combustion process simulator, which serves as an approximated dynamics model $f(s_t, a_t)=\hat{s}_{t+1}, \hat{r}_{t+1}, \hat{c}_{t+1}$ to generate future states, rewards and costs in our model-based RL framework. 

Accurately fitting the dynamics of combustion process is very challenging. TPGUs are highly complex, large, and partially observed open systems. Due to extremely high temperature and pressure in certain parts of a TPGU, some state information is not fully captured by sensors. External factors like ambient temperature and chemical properties of the coal also impact combustion. 
We propose a customized deep recurrent neural network (RNN) as the combustion process simulator,  with its internal cell structure specially designed according to the actual physical process. As shown in Figure \ref{fig.simulator}, the input state-action pairs are split into 3 blocks to encode their physical and hierarchical dependencies, and the long short term memory (LSTM) layers are used to capture the temporal correlations.
Specifically, we first model the coal pulverizing stage by considering the related states $s_t^c$ and actions $a_t^c$ together with the external inputs $s_t^e$ (e.g. environment temperature and chemical properties of the coal), and predict the next coal pulverizing related states $s_{t+1}^c$. We then combine the impact from the coal pulverizing stage (encoded in the hidden states $h_t^c$) with the states $s_t^b$ and actions $a_t^b$ of burning stage to predict the next state $s_{t+1}^b$. Lastly, impacts of burning stage $h_t^b$ are combined with the states $s_t^s$ and actions $a_t^s$ of steam circulation stage to predict the related states of next time step $s_{t+1}^s$. This design embeds domain knowledge in the network structure, which helps to alleviate the impact of missing information in the partially observed system, and greatly improves model accuracy and robustness.

The simulator is learned by minimizing the mean squared error of the actual and predicted states.
To further strengthen the simulator, following techniques are applied: 1) \textbf{Seq2seq} and \textbf{scheduled sampling}: We use sequence to sequence structure and scheduled sampling \cite{bengio2015scheduled} to improve long-term prediction accuracy. 2) \textbf{Noisy data augmentation}: We add gradually vanishing Gaussian noises on the state inputs during training, which can be perceived as a means of data augmentation. This helps to improve model robustness and prevent overfitting.

\section{MORE: An Improved Model-Based Offline RL Framework}

\begin{figure}[t]
	\includegraphics[width=0.4\textwidth]{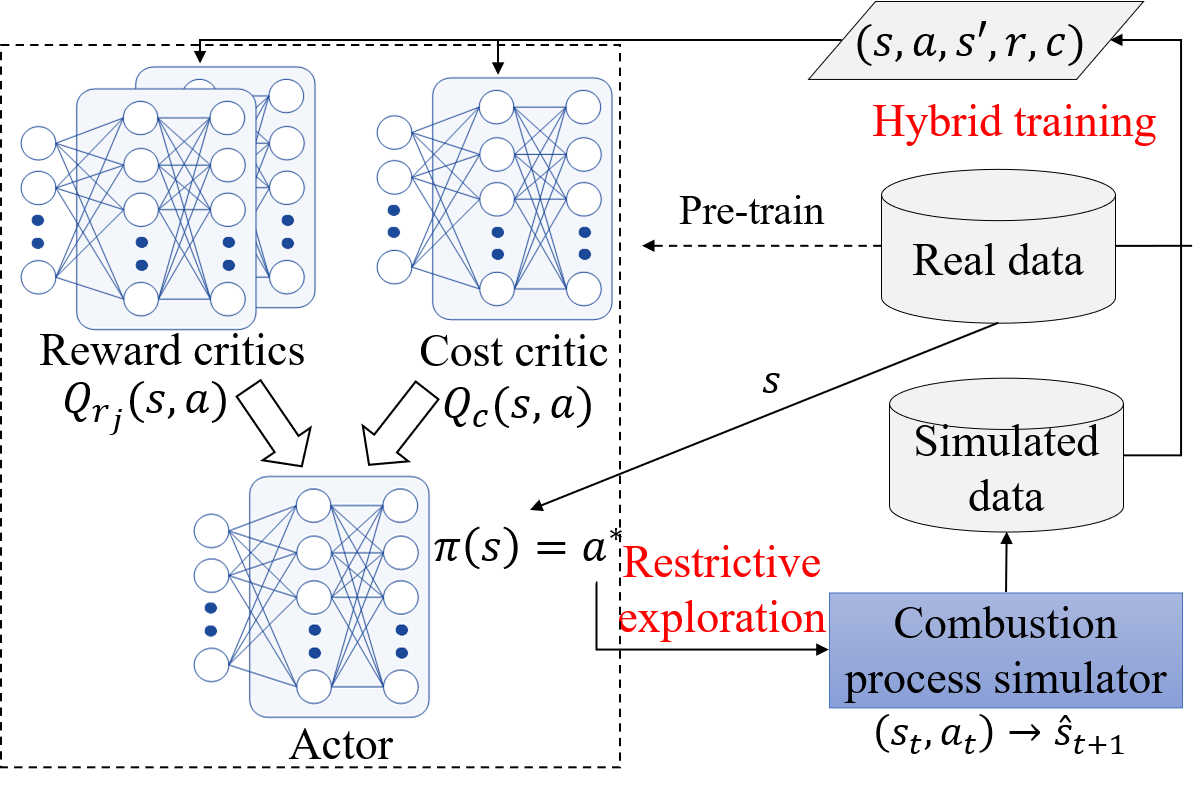}
	\centering
	\caption{The framework of MORE}
	\label{fig:more_framework}
\end{figure}

In this Section, we introduce the core RL algorithm used in DeepThermal: \underline{M}odel-based \underline{O}ffline RL with \underline{R}estrictive \underline{E}xploration (MORE).
MORE tackles the challenge of offline policy learning under constraints with an imperfect simulator. The framework of MORE is illustrated in Figure \ref{fig:more_framework}. 
It introduces an additional cost critic to model and enforces safety constraints satisfaction of the combustion optimization problem. MORE quantifies the risks imposed by the imperfect simulator using a novel \textit{restrictive exploration} scheme, 
from the perspective of both prediction reliability (measured by \textit{model sensitivity}) as well as the possibility of being OOD samples (measured by \textit{data density} in the behavioral data). 
Specifically, MORE trusts the simulator only when it is certain about the outputs and adds reward penalty on potential OOD predictions to further guide the actor to explore in high density regions. Finally, MORE ingeniously combines the real data and carefully distinguished simulated data to learn a safe policy through a \textit{hybrid training} procedure.

\subsection{Safe Policy Optimization}

MORE uses 
two types of Q-functions, $Q_r$ and $Q_c$, for reward maximization and cost evaluation. The policy optimization is performed on the carefully combined real-simulation data $\mathcal{D}$:
\begin{equation}
\label{eq:opt}
\begin{split}
\pi_{\theta}:=\max_{\pi} &~\mathbb{E}_{s\sim\mathcal{D}, a\sim\pi}\left[\min_{j=1,2} Q_{r_{j}}(s, a)\right] \\
&\text{ s.t. } \mathbb{E}_{a\sim\pi}[Q_c(s, a)] \leq l
\end{split}
\end{equation}
MORE adopts the Clipped Double-Q technique \cite{fujimoto2018addressing} by using two $Q_r$ functions to penalize the uncertainty in $Q_r$ and alleviate the overestimation issue that commonly occurs in off-policy RL. This trick is not applied to the $Q_c$-network as it could potentially underestimate the cost value.
To solve this problem, we employ the Lagrangian relaxation procedure \cite{boyd2004convex} to convert the original problem (Eq. \ref{eq:opt}) to the following unconstrained form:
\begin{align}
\mathcal{L}(\pi, \lambda)=&\mathbb{E}_{s \sim \mathcal{D}, a\sim\pi}\Big[\min_{j=1,2} Q_{r_{j}}(s, a) -\lambda \left(Q_c(s, a) - l\right) \Big] \notag\\\
&\left(\pi^{*}, \lambda^{*}\right)=\arg \min _{\lambda \geq 0} \max _{\pi} \mathcal{L}(\pi, \lambda). \label{eq:unconstrained}\\
&\lambda\leftarrow[\lambda+\eta(\mathbb{E}_{a\sim\pi}[Q_c(s, a)]-l)]^{+} \label{eq:update_lambda}
\end{align}
where $\lambda$ is the Lagrangian multiplier, $\eta$ is the step size and $[x]^{+}=\max \{0, x\}$. We use the iterative primal-dual update method to solve the above unconstrained minimax problem. In the primal stage, we fix the dual variable $\lambda$ and perform policy gradient update on policy $\pi$.
In the dual stage, we fix the policy $\pi$ and update $\lambda$ by dual gradient ascent (Eq. \ref{eq:update_lambda}).

\subsection{Restrictive Exploration}

Like many real-world tasks, we only have an imperfect simulator for the combustion optimization problem. The model is learned entirely from the offline data and possible to make inaccurate predictions that impact RL training. The inaccuracies are mainly from two sources: 1) lack of data to fully train the model in certain regions; 2) unable to well fit data due to limited model capability or system complexity. Note the first case is not an absolute criteria to detect model inaccuracy. Models can perform reasonably well in low density regions of data if the pattern is easy to learn. Under this setting, we should encourage exploration with the model even if the resulting samples lie outside the dataset distribution.

With this intuition, we design a new \textit{restrictive exploration} strategy to fully utilize the generalizability of the simulator from both the model and data perspective. The key insight is to only consider the samples that the simulator is certain, and then further distinguish whether the simulated samples are in data distribution or not.

\noindent \textbf{Model sensitivity based filtering.}
We first filter out those unreliable simulated data if the model is uncertain. Previous work \cite{novak2018sensitivity} has shown that model sensitivity can be a viable measure for model generalizability on data samples. We use this metric to detect if the model is certain or well generalizable on simulated state-action pairs from the model's perspective.
For sensitivity quantification, we inject $K$ i.i.d. Gaussian noises $\epsilon_i \sim N(\mathbf{0},\sigma \mathbf{I})$, $i\in\{1,\dots,K\}$ on a input state-action pair $(s,a)$ of the model, and compute the variance of the output perturbations $u=Var[\boldsymbol{\epsilon}^y]$ as the sensitivity metric, where $\boldsymbol{\epsilon}^y=\big[f((s,a)+\epsilon_i)-f(s,a)\big]_{i=1}^K$, and $\sigma$ controls the degree of perturbation. A large $u$ suggests the model is sensitive to input perturbation at $(s, a)$, which is an indication of uncertainty or lack of prediction robustness at this point \cite{novak2018sensitivity}.

Let $\tau_s$ be a batch of simulated transitions $\{(s,a,s',r,\tilde{c})\}$ at training step $t$ and $\mathbf{u}_{s,t}$ be the sensitivity of $\tau_s$. MORE filters problematic simulated transitions as follows:
\begin{equation}
\label{eq:filter}
\begin{split}
\tau_m:=\{ \tau_s|\mathbf{u}_{s,t}<l_u\}
\end{split}    
\end{equation}
where $l_u$ is a predefined threshold. In practice, we choose it to be the $\beta_u$-percentile value of sensitivity pre-evaluated at all state-action pairs in the offline dataset $\mathcal{B}$.

\noindent \textbf{Data density based filtering.}
The lack of data in low density regions of $\mathcal{B}$ may provide insufficient information to describe the system dynamics completely and accurately. This can lead to unreliable OOD simulated transitions that cause exploitation error during policy learning.
To address this issue, we propose the data-density based filtering to encourage exploration in high density regions, while cautioning about potential OOD samples. The key insight is to carefully distinguish between positive (in high density region) and negative (in low density or OOD) simulated samples. We trust more on positive samples while penalizing on the negative samples. 

In practical implementation, we use a state-action variational autoencoder (VAE) \cite{kingma2014auto} to fit the data distribution of $\mathcal{B}$. VAE maximize the following evidence lower bound (ELBO) objective that lower bounds the actual log probability density of data:
\begin{equation}
\label{eq:cal_pm}
\mathbb{E}_{z \sim q_{\omega_2}}\left[\log p_{\omega_1}(s,a|s,a,z)\right] - D_{\text{KL}}\left[q_{\omega_2}(z|s,a) \| N(0,1)\right]
\end{equation}
where the first term represents the reconstruction loss and the second term is the KL-divergence between the encoder output and the prior $N(0,1)$.
We use ELBO to approximate the probability density of data.
Let $\tau_m$ be the simulation samples that passed model sensitivity based filtering
at training step $t$. We estimate data density $p_m$ of state-action pairs in $\tau_m$ with Eq. \ref{eq:cal_pm} and split $\tau_m$ to positive samples $\tau_+$ and negative samples $\tau_-$ with threshold $l_p$.
\begin{equation}
\label{eq:split}
\begin{split}
\tau^+:=\{\tau_m|p_m>l_p\},\quad \tau^-:=\{\tau_m|p_m\leq l_p\}
\end{split}
\end{equation}
Like $l_u$, we choose $l_p$ to be the $\beta_p$-percentile ELBO values pre-evaluated on all state-action pairs in the dataset $\mathcal{B}$.
\begin{algorithm}[t]
	\caption{Restrictive exploration}
	\label{alg:restrictive}
	\begin{algorithmic}[1]
		\STATE \textbf{Require:} Simulator $f$, threshold 
		$l_u$, $l_p$,
		batch of real data transitions $\tau_n=\{(s, a, s', r, \tilde{c})\}_n$, and rollout length $H$
		\FOR{$\tau$ in $\tau_n$}
		\STATE Set $\hat{s}_1 = s$, $\tau^+=\emptyset$, $\tau^-=\emptyset$
		\FOR{Rollout step: $h=1,..,H$}
		\STATE Generate transition 
		$(\hat{s}_{h}, \pi(\hat{s}_{h}), \hat{s}_{h+1}, \hat{r}_{h}, \hat{c}_{h})$, 
		where $(\hat{s}_{h+1}, \hat{r}_{h}, \hat{c}_{h})=f(\hat{s}_{h},\pi(\hat{s}_{h}))$
		\STATE \textbf{// Model sensitivity based filtering}\\
		\IF{Sensitivity $u(\hat{s}_h,\hat{a}_h)<l_u$ (follow Eq.\ref{eq:filter})}
		\STATE \textbf{// Data density based filtering} \\
		Compute data density $p_m(\hat{s}_h,\hat{a}_h)$ as in Eq.\ref{eq:cal_pm}
		\STATE Add $(\hat{s}_{h}, \pi(\hat{s}_{h}), \hat{s}_{h+1}, \hat{r}_{h}, \hat{c}_{h})$ into positive sample set $\tau^+$ or negative set $\tau^-$ according to Eq.\ref{eq:split}
		\STATE Add reward penalties for $\tau^-$ samples as in Eq.\ref{eq:replace_r}
		\ENDIF
		\ENDFOR
		\ENDFOR
		\STATE \textbf{Output:}  ($\tau^+, \tau^-$)
	\end{algorithmic}
\end{algorithm}

\subsection{Hybrid Training}

After quantifying the risks imposed by the imperfect simulator, MORE introduces a hybrid training strategy to differentiate the impact of positive and negative simulated samples obtained from the restrictive exploration.
We keep the positive samples as their original forms to encourage fully exploiting the generalizability of the model, but penalize the rewards of negative samples to guide policy learning away from high-risk regions.
Moreover, we pretrain $\pi$, $Q_r$ and $Q_c$ with real data in order to run RL algorithm with good initial parameters, which is observed to improve stability of training and speed up convergence.

\noindent \textbf{Reward penalization on negative samples.}
Several previous works in online and offline RL \cite{yu2020mopo,kidambi2020morel,shi2019virtual} already use penalized rewards to regularize policy optimization against potential negative impacts of simulated samples. 
However, unlike prior works that penalize the reward of all simulated samples, we use a more delicate strategy by softly penalizing rewards as:
\begin{equation}
\label{eq:replace_r}
\begin{split}
\hat{r}(\hat{s}_t,\hat{a}_t)\leftarrow\frac{\hat{r}(\hat{s}_t,\hat{a}_t)}{1+[\kappa(l_p-p_m(\hat{s}_t,\hat{a}_t))]^+}
\end{split}
\end{equation}
where 
$\kappa$ is a hyper-parameter to control the scale of reward penalty. It's easy to find that positive samples whose approximated density $p_m(\hat{s}_t,\hat{a}_t)$ higher than $l_p$ are not penalized (see Eq.\ref{eq:split}). Only negative samples are penalized and the penalty weight is propotional to the difference between $p_m(\hat{s}_t,\hat{a}_t)$ and $l_p$. This strategy encourages policy updates toward high reward directions suggested by positive samples, providing the possibility to generalize beyond the offline dataset; while also forcing policy updates away from the area of OOD negative samples to avoid potential exploitation error.

MORE constructs a special local buffer $\mathcal{R}$ to combine real, positive and negative simulated data for offline training.
The full algorithm is summarized in Algorithm \ref{alg:more}. 

\begin{algorithm}[t]
	\caption{Complete algorithm of MORE}
	\label{alg:more}
	\begin{algorithmic}[1]
		\STATE \textbf{Require:} Offline dataset $\mathcal{B}$
		\STATE Pre-train actor $\pi_\theta$, reward critic ensemble $\{Q_{r_i}(s,a|\phi_{r_i})\}_{i=1,2}$ and cost critic $Q_c(s,a|\phi_{c})$ with real data. Initialize target networks $\{ Q'_{r_i} \}_{i=1}^{2}$ and $Q'_{c}$ with $\phi'_{r_i} \leftarrow \phi_{r_i}$ and $\phi'_{c} \leftarrow \phi_{c}$
		\FOR{Training step: $t=1,...,T$}
		\STATE Random sample mini-batch transitions $\tau_n$ from $\mathcal{B}$
		\STATE Obtain $(\tau^+, \tau^-)$ using restrictive exploration (Alg. \ref{alg:restrictive})
		\STATE Construct local buffer $\mathcal{R}=\{(s,a,r,c,s')\}$ using $\tau^+, \tau^-$ and $\tau_n$, as well as Eq.\ref{eq:replace_r}
		\STATE Set $y=\min_{i=1,2} Q'_{r_{i}}\left(s', \pi(s')\right)$, $z= Q'_{c}\left(s', \pi(s')\right)$
		\STATE Update $Q_{r_i}$ by minimizing $(Q_{r_i}-(r+\gamma y))^{2}$ 
		\STATE Update $Q_{c}$ by minimizing $(Q_{c}-(c+\gamma z))^{2}$
		\STATE Update policy $\pi_\theta$ by Eq.\ref{eq:unconstrained} using policy gradient 
		\STATE Update $\lambda$ by Eq.\ref{eq:update_lambda} using dual gradient ascent 
		\STATE Update target cost critic: $\phi'_{c} \leftarrow \rho \phi_{c}+(1-\rho)\phi'_{c}$
		\STATE Update target reward critics: $\phi'_{r_i} \leftarrow \rho \phi_{r_i}+(1-\rho)\phi'_{r_i}$
		\ENDFOR
	\end{algorithmic}
\end{algorithm}

\section{Experiments}


\subsection{Dataset and Settings}

We conduct experiments on both real TPGUs and standard offline RL benchmarks. 
Detailed settings are as follows:

\noindent \textbf{Real-world datasets and experiment settings.}
We used 1$\sim$2 years' historical TPGU operational data to train our models. Very old data are not used due to potentially different patterns compared with current conditions of the TPGU, mainly caused by changes and deteriorating of equipment and devices.
We considered more than 800 sensors and optimized about 100 control variables. A specially designed feature engineering process is used to process these sensor data into about 100$\sim$170 states and $30\sim50$ actions (differ for TPGUs in different power plants). Some sensor values monitoring similar state as well as control variables sharing the same operation mode are merged into single values to reduce problem dimension.
Finally, we re-sampled the processed data into equal 20$\sim$30 second (depending on the quality of the sensor data) interval data, which typically results in 1$\sim$2 million records for RL training.


\noindent \textbf{Datasets and settings for standard offline RL benchmarks.}
We evaluate and compare the performance of MORE on the standard offline RL benchmark D4RL \cite{fu2020d4rl}.
 We mainly focus on three locomotion tasks (hopper, halfcheetah and walker2d) and two dataset types (medium and mixed) that are more relevant to real-world applications.
These datasets are generated as follows: \textbf{medium}: generated using a partially trained SAC policy to roll out 1 million steps. \textbf{mixed}: train a SAC policy \cite{haarnoja2018soft} until reaching a predefined performance threshold, and take the replay buffer as the dataset.
For all experiments on D4RL datasets, we model the dynamics model using fully connected neural networks.

\subsection{Evaluation of the Simulator}

We compare in Table \ref{tab.sim_eval} the performance of the combustion process simulator with four baselines that commonly used for time-series data prediction, including ARIMA, GBRT, DNN (feedforward neural network) and stacked LSTM. 
It is observed that our proposed combustion process simulator significantly outperforms all the baselines on both evaluation metrics (RMSE and MAE). This demonstrates the effectiveness of the proposed simulator, as well as the benefit of incorporating domain knowledge in the network design.

\begin{table}[b]
	\small
	\centering
	
	\begin{tabular}{ c|c|c|c|c|c} 
		\hline
		\textbf{Model}&\textbf{ARIMA}&\textbf{GBRT}&\textbf{DNN}&\textbf{LSTM}&\textbf{Ours}\\
		\hline
		\textbf{RMSE}&3.05e-1&1.97e-1&2.05e-2&1.69e-3&\textbf{6.54e-4}\\
		\hline
		\textbf{MAE}&2.66e-1&2.65e-1&2.73e-2&2.50e-2&\textbf{1.55e-3}\\
		\hline
	\end{tabular}
	\caption{Evaluation of the combustion process simulator}
	\label{tab.sim_eval}
\end{table}

\begin{figure*}[tbp]
	\centering
	\subfigure[270 MW Experiment]{
		\includegraphics[width=0.32\textwidth]{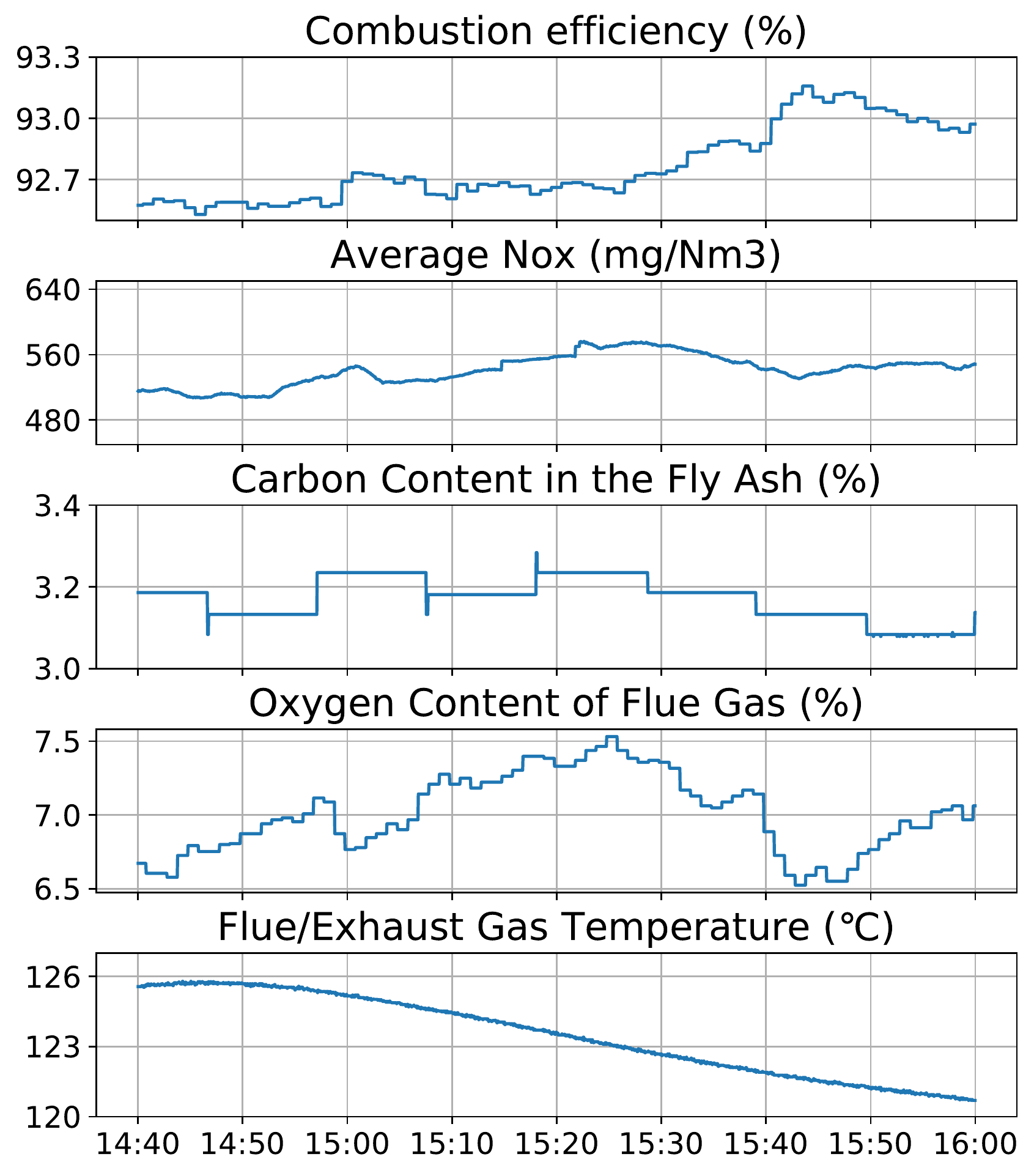}
	}
	\subfigure[290MW Experiment]{
		\includegraphics[width=0.32\textwidth]{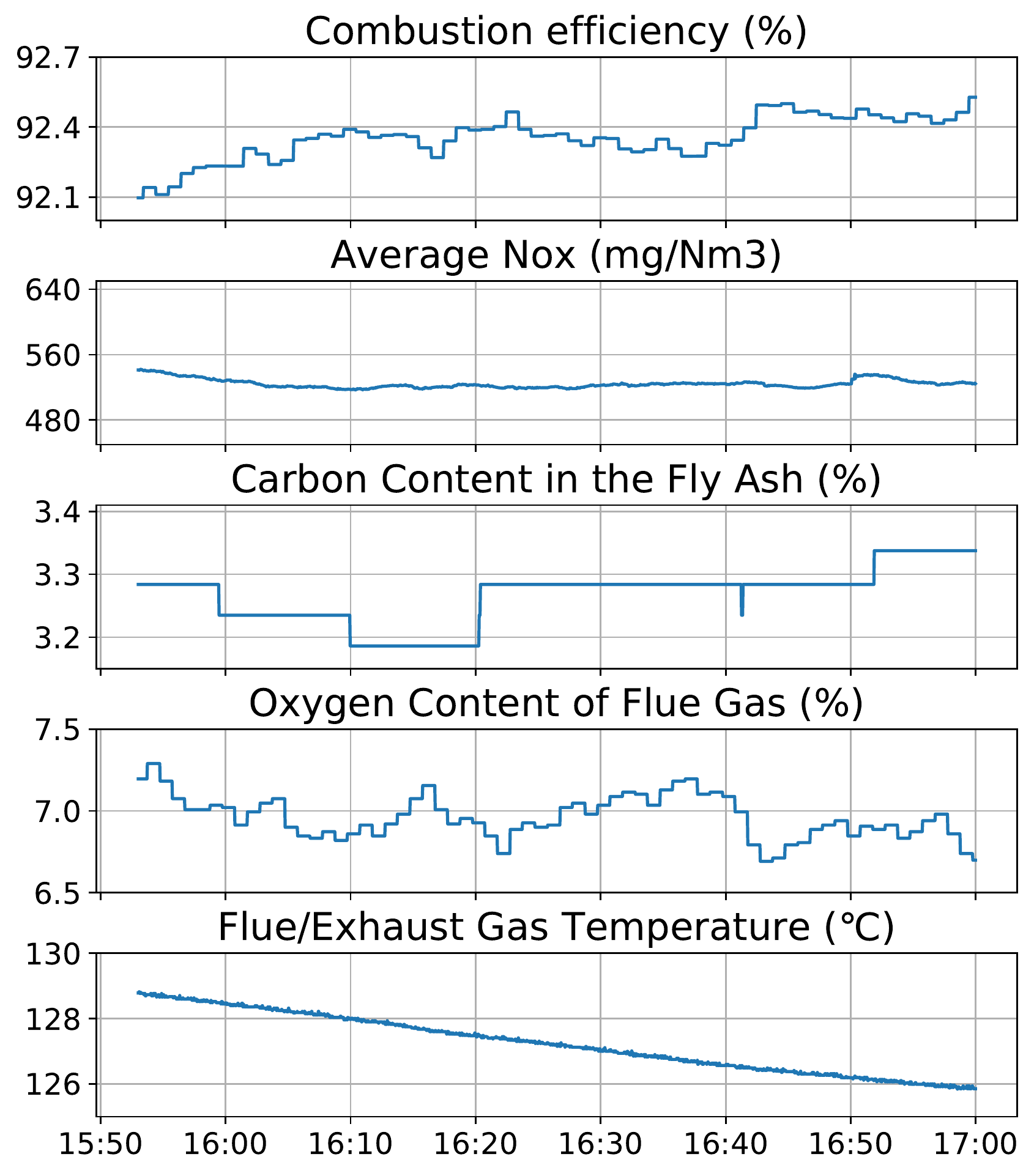}
	}
	\subfigure[310MW Experiment]{
		\includegraphics[width=0.32\textwidth]{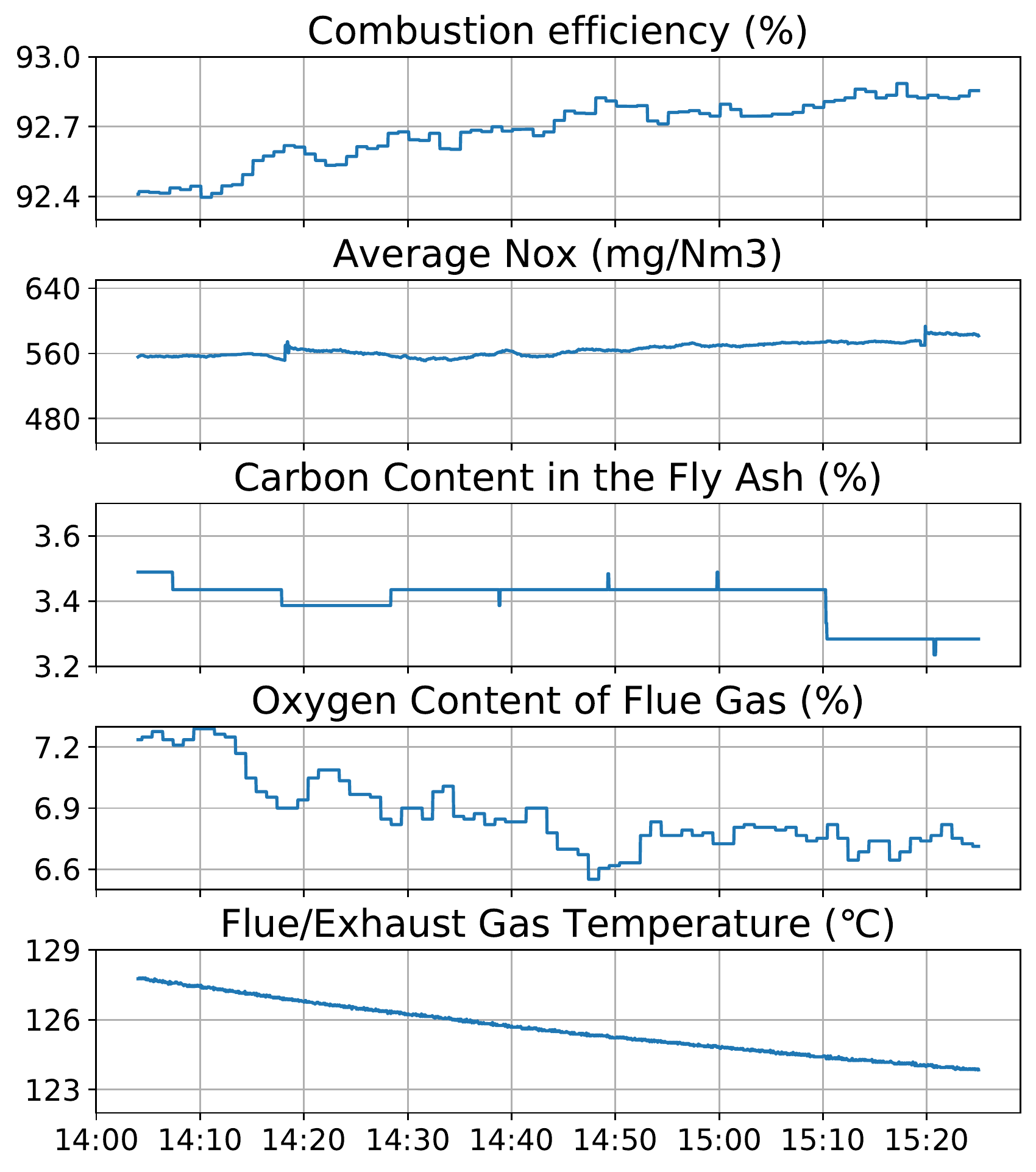}
	}
	\caption{Real-world experiments at CHN Energy Nanning Power Station}
	\label{fig:real_exp} 
\end{figure*}

\begin{table*}[htbp]
	\small
	\centering
	\begin{tabular}{c|cc|cccccc|c}
		\hline
		\textbf{Dataset}   & \textbf{Batch Mean} & \textbf{Batch Max} & \textbf{BC} & \textbf{BEAR} & \textbf{BRAC-v} & \textbf{BCQ} & \textbf{MBPO} & \textbf{MOPO} & \textbf{MORE (Ours)} \\ \hline
		halfcheetah-medium      & 3953                & 4410.7             & 4202.7      & 4513.0        & 5369.5          & 4767.9       & 3234.4 & 4972.3  & \textbf{5970}   \\
		hopper-medium & 1021.7              & 3254.3             & 924.1       & \textbf{1674.5}        & 1031.4          & \textbf{1752.4}          & 139.9   & 891.5   & 1264   \\
		walker2d-medium    & 498.4               & 3752.7             & 302.6       & 2717.0        & \textbf{3733.4}          & 2441.3       & 582.8   & 817.0   & \textbf{3649} \\ \hline
		halfcheetah-mixed       & 2300.6              & 4834.2             & 4488.2      & 4215.1        & 5419.2          & 4463.9       & 5593.0 & \textbf{6313.0}  & 5790   \\
		hopper-mixed  & 470.5               & 1377.9             & 364.4       & 331.9         & 9.7             & 688.7        & 1600.8  & \textbf{2176.8}  & \textbf{2100}   \\
		walker2d-mixed     & 358.4               & 1956.5             & 518.5       & 1161.4        & 36.2            & 1057.8       & 1019.1  & 1790.7  & \textbf{1947}   \\ \hline
	\end{tabular}
	\caption{Results for D4RL datasets, averaged over 5 random seeds}
	\label{tab:result}
\end{table*}

\subsection{Real-World Experiments}

To verify the effectiveness of DeepThermal, we conducted a series of before-and-after tests on real-world TPGUs. The duration of these experiments ranges from 1 to 1.5 hours.
During the experiment, the human operator adjusted the control strategy of a TPGU according to the recommended actions provided by the learned RL policy. 

Figure \ref{fig:real_exp} reports the experiment results in CHN Energy Nanning Power Station on three different load settings (270MW, 290MW, 310MW).
It is observed that DeepThermal effectively improves combustion in all three load settings, with the 
maximum increase of 0.52\%, 0.31\% and 0.48\% on the combustion efficiency in about 60 min compared with the initial values. The average $\mathrm{NO_x}$ concentrations before denitrification reactor remain at a relative stable level. We also present three key indicators that reflect sufficient combustion, including \textit{carbon content in the fly ash}, \textit{oxygen content of flue gas} and \textit{flue/exhaust gas temperature}. 
In all experiments, these three indicators achieved a certain level of decrease, providing clear evidences of combustion improvement.


\subsection{Evaluation on Offline RL Benchmarks}
We further investigate the performance of our model-based offline RL framework MORE on the D4RL benchmarks.

\begin{figure}[t]
	\centering
	\subfigure[Ablation of $\beta_u$] {
		\includegraphics[width=0.47\columnwidth]{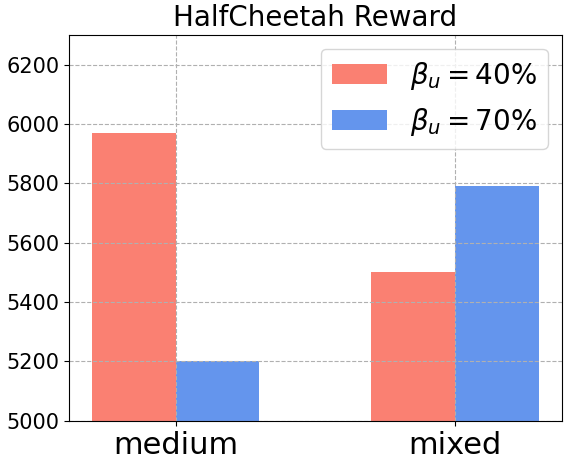} 
	}
	\subfigure[Ablation of $\beta_p$] {
		\includegraphics[width=0.47\columnwidth]{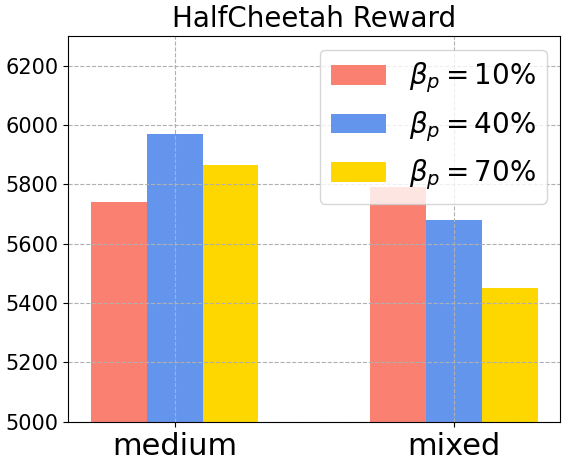} 
	}
	\subfigure[Evaluation of different safety constraints $l$] {
		\includegraphics[width=\columnwidth]{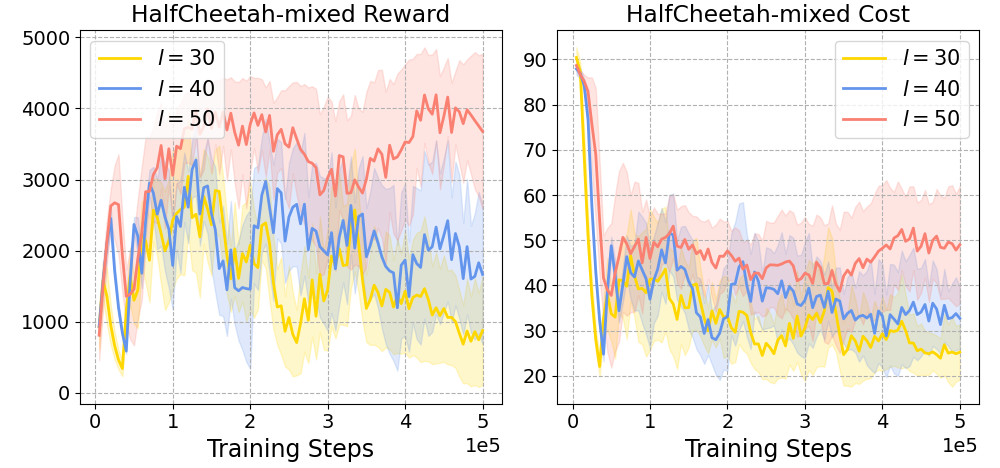} 
	}
	\caption{Ablation study on MORE}
	\label{ablation}
\end{figure}

\noindent \textbf{Comparative Evaluations.}
We compare MORE against the state-of-the-art offline RL algorithms, including model-free algorithms such as BCQ \cite{fujimoto2019off}, BEAR \cite{kumar2019stabilizing} and BRAC-v \cite{wu2019behavior} that constrain policy learning to stay close to the behavior policy using various divergence metrics. We also compare against model-based offline RL algorithms including MOPO \cite{yu2020mopo} that follows MBPO \cite{janner2019trust} with additional reward penalties.
We omit the cost critic of MORE in these experiments, as there are no safety constraints in corresponding D4RL tasks.

The results are presented in Table \ref{tab:result}. 
It is observed that MORE matches or outperforms both the model-free and model-based baselines in most tasks. MOPO is shown to outperform model-free methods by a large margin in the mixed datasets, while performs less well on the medium datasets due to the lack of action diversity. MORE matches the performance of MOPO on the mixed datasets while greatly surpasses MOPO on the medium datasets. We hypothesize that conditionally removing uncertain simulated samples (via model sensitivity based filtering) as well as introducing data-density based reward penalties on OOD samples provide more reliable and informative simulated data for policy learning, thus lead to good results with an imperfect model.

\noindent \textbf{Ablation Study}
We conduct a series of ablation studies on halfcheetah environment to investigate how different components impact the performance of MORE.

\begin{itemize}[leftmargin=*, topsep=0pt]
	\item \textit{Evaluation on the model sensitivity threshold $\beta_u$.} It can be shown in Figure \ref{ablation}(a) that, in the mixed dataset where the simulator can learn and generalize well, MORE with $\beta_u=70\%$ outperforms $\beta_u=40\%$ (more tolerant to encourage generalization). While in the medium dataset, MORE with $\beta_u=70\%$ performs inferior than 
	$\beta_u=40\%$ due to allowing too much problematic samples from the imperfect dynamics models. 
	\item \textit{Evaluation on the data density threshold $\beta_p$.} 
	We find in Figure \ref{ablation}(b) that smaller $\beta_p$ ($\beta_p=10\%$) performs better in the mixed dataset, while a medium value $\beta_p$ ($\beta_p=40\%$) works better in the medium dataset. This again suggests that it is beneficial to 
	be more tolerant to potential OOD simulated samples 
	when the dynamics model is reliable. However, when the dynamics model is imperfect, carefully controlling the ratio between positive and negative samples is important to achieve the best performance.
\end{itemize}

\noindent \textbf{Additional Evaluation under Safety Constraints.}	
We conduct additional experiments to demonstrate the performance of MORE under safety constraints. We use the halfcheetah-mixed dataset and further introduce the safety cost as the discounted cumulative torque that the agent has applied to each joint \cite{tessler2018reward}.
The per-state cost $c(s, a)$ is the amount of torque the agent decided to apply at each step, i.e. the L2-norm of the action vector, $\|a\|_{2}$.
It should be noted that by preventing the agent from using high torque values, the agent may learn a sub-optimal policy. 
We test MORE under different constraint limit $l \in \{30, 40, 50 \}$. It can be shown in Figure \ref{ablation}(c) that MORE is robust to different $l$. In all the tests, the cumulative costs are controlled below the given constraint limits.

\section{Related Work}

\subsection{Complex System Control}
PID control \cite{astrom2006advanced} is the most common approach for industrial system control. Although PID control ensures safe and stable control, its performance is limited due to insufficient expressive power. Model predictive control (MPC) \cite{garcia1989model} is another widely used control method, that utilizes an explicit process model to predict the future response of the system and performs control optimization accordingly. MPC has been applied to many areas, such as refining, petrochemicals, food processing, mining/metallurgy and automotive applications \cite{qin2003survey}. 
However, applying MPC in large-scale stochastic systems is often infeasible due to their heavy online computational requirements. 

RL overcomes the above challenges by learning the optimal strategy beforehand, a concept similar to parametric programming in explicit model predictive control \cite{bemporad2002explicit}.
Previous works that use RL for real-world control tasks typically rely on existed high-fidelity simulators \cite{li2019reinforcement,lopez2018microscopic,todorov2012mujoco}, or build "virtual" simulators using diverse and large data \cite{shi2019virtual,lazic2018data}.
However, both high-fidelity simulators and diverse data are impossible to obtain in some complex real-world tasks, using data-driven RL algorithms hold the promise of automated decision-making informed only by logged data, thus getting rid of the sim-to-real dilemma \cite{dulac2020empirical}.

\subsection{Offline Reinforcement Learning}
Offline RL focuses on learning policies from offline data without environment interaction.
One major challenge of offline RL is distributional shift \cite{levine2020offline}, which incurs when policy distribution deviates largely from data distribution. 
Recent model-free methods attempted to solve this problem by constraining the learned policy to be close to the behavior policy via implicit or explicit divergence regularization \cite{fujimoto2019off,kumar2019stabilizing,wu2019behavior,kumar2020conservative,xu2021constraints,xu2021offline}. 
While performing well in single-modal datasets, model-free methods are shown to have limited improvements in multi-modal datasets due to over-restricted constraints.

Model-based RL algorithms provide another solution to offline RL. They adopt a pessimistic MDP framework \cite{kidambi2020morel}, where the reward is penalized if the learned dynamic model cannot make an accurate prediction.
MOPO \cite{yu2020mopo} extends MBPO \cite{janner2019trust} with an additional reward penalty on generated transitions with large variance from the learned dynamic model. 
MOReL \cite{kidambi2020morel} terminates the generated trajectories if the state-action pairs are detected to be unreliable, i.e. the disagreement within model ensembles is large.
MBOP and MOPP \cite{argenson2021modelbased,zhan2021model} learn a dynamics model, a behavior policy and a value function to perform model-based offline planning, where the actions are sampled from the learned behavior policy.
Note that all these methods largely depend on the quality of learned dynamic models, while MORE reduces the reliance on the model by using information from both the model and offline data.

\section{Conclusion}
We develop DeepThermal, a data-driven AI system for optimizing the combustion control strategy for TPGUs. To the best of the authors' knowledge, DeepThermal is the first offline RL application that has been deployed to solve real-world mission-critical control tasks.
The core of DeepThermal is a new model-based offline RL framework, called MORE. MORE strikes the balance between fully utilizing the generalizability of an imperfect model and avoiding exploitation error on OOD samples.
DeepThermal has been successfully deployed in four large coal-fired power plants in China. 
Real-world experiments show that DeepThermal effectively improves the combustion efficiency of TPGUs.
We also conduct extensive comparative experiments on standard offline RL benchmarks to demonstrate the superior performance of MORE against the state-of-the-art offline RL algorithms.



\section{Acknowledgments}
A preliminary version of this work was accepted as a spotlight paper in RL4RealLife workshop at ICML 2021. This work was partially supported by the National Key R\&D Program of China (2019YFB2103201).

\bibliography{aaai22}

\newpage
\clearpage
\appendix

\section{Appendix}
\subsection{Overall System Framework}

\begin{figure}[h]
	\centering
	\includegraphics[width=1.05\columnwidth]{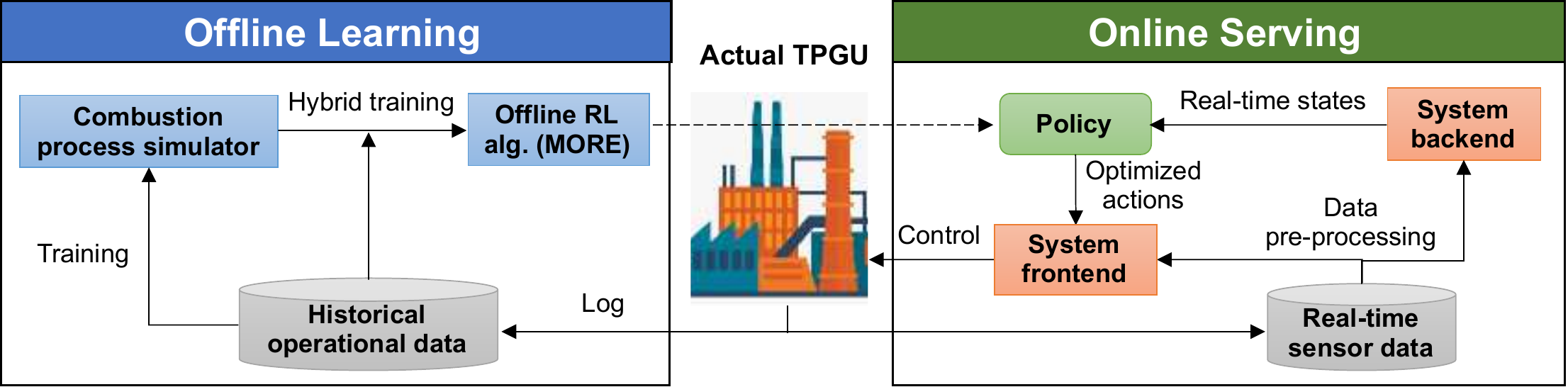}
	\caption{Overall framework of the DeepThermal system}\label{fig:deepthermal}
\end{figure}

DeepThermal consists of two parts: \textit{offline learning} and \textit{online serving}, which are illustrated in Figure \ref{fig:deepthermal}.	
Due to the complexity of the combustion process in a TPGU, it is impossible to build a high-fidelity simulation environment. 
Solely using 1 or 2 years' operational data may not be sufficient to find the optimized control strategy. 
In the offline learning part, DeepThermal learns a data-driven simulator and adopts a model-based offline learning framework (MORE) to combat the limited data issue. The combustion simulator is used to provide supplement dynamics data to facilitate RL training. It is also used to generalize beyond the existing stereotyped control strategies of human operators recorded in data. However, as the simulator is learned from data, we do not fully trust the simulated data and use them with extra caution. Inside MORE, we introduce several specially designed strategies, including restrictive exploration and hybrid training to filter problematic simulated data and introduce reward penalties on OOD samples to guide offline RL policy learning away from high-risk areas.

During online serving, the learned RL policy outputs optimized actions according to the state processed from real-time sensor data streams by the system backend.
The system frontend displays the optimized control strategies to human operators, who adjust the combustion control of a TPGU.

As conditions of the equipment and devices inside a TPGU can change or deteriorate over time. DeepThermal is designed to be completely data-driven, which allows  re-collecting the newly generated operational data from the TPGU for RL policy re-training and fine-tuning every few months.
After every few months, we can re-collect the newly generated operational data of a TPGU to finetune the existing RL policy. 
This enables the models in DeepThermal to adapt to the current condition of the TPGU, providing an evolving optimization solution to a slowly changing system.

\subsection{Real-World System Deployment}

DeepThermal has already been successfully deployed in four large coal-fired thermal power plants in China, including CHN Energy Nanning and Langfang Power Stations, 
Shanxi Xingneng Power Station and Huadian Xinzhou Guangyu Power Station.
Real-world experiments have been conducted in all four power plants to test the effectiveness of DeepThermal. 
Our system achieves good results while ensures safe operation in all these four power plants, and has passed the project acceptance checks by industry experts, whom consider highly on the innovation as well as the effectiveness of our systems.

\begin{figure*}[ht]
	\centering
	\includegraphics[width=0.9\textwidth]{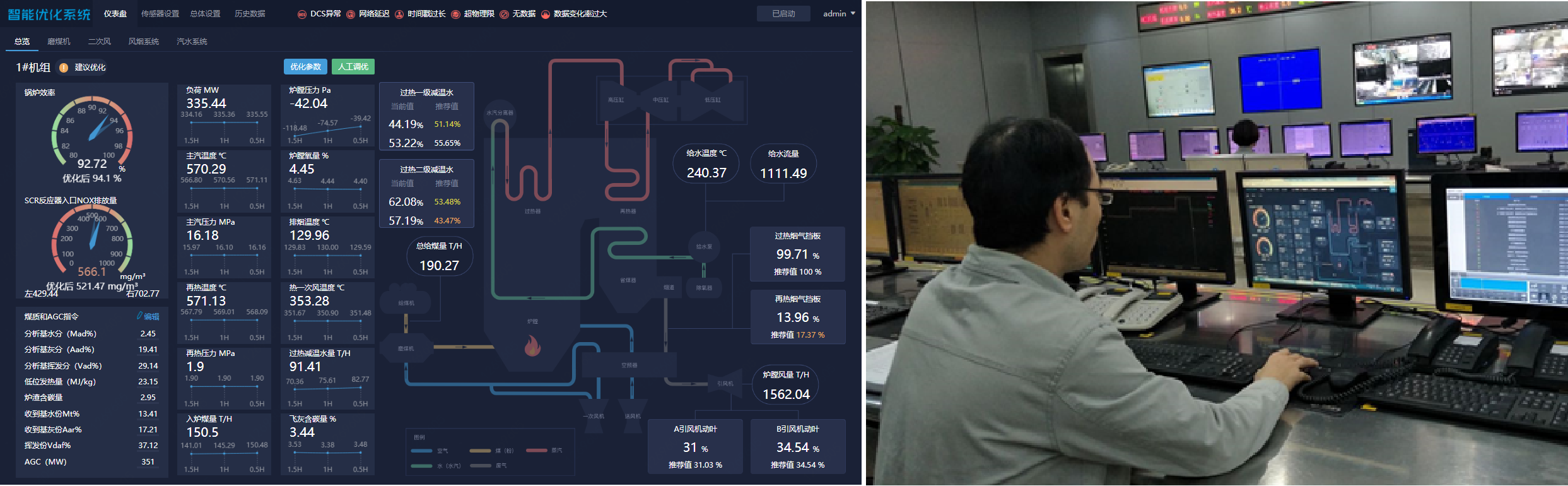}
	\caption{Interface of DeepThermal deployed in CHN Energy Nanning Power Station (left) and the control room (right)}
	\label{sys.dep}
\end{figure*}

The left figure in Figure \ref{sys.dep} shows the main interface of the DeepThermal system deployed in CHN Energy Nanning Power Station. The interface displays real-time values of major states as well as optimized actions provided by the learned RL policy.
The operator can easily follow the guidance of the recommended strategy to adjust their control operation, so as to improve combustion efficiency of the TPGU.
The right figure in Figure \ref{sys.dep} shows the scene that the operator of the power plant using DeepThermal for reference to adjust the combustion control in the central control room. 

\begin{figure*}[ht]
	\centering
	\subfigure[System deployed in CHN Energy Langfang Power Station]{
		\includegraphics[width=0.9\textwidth]{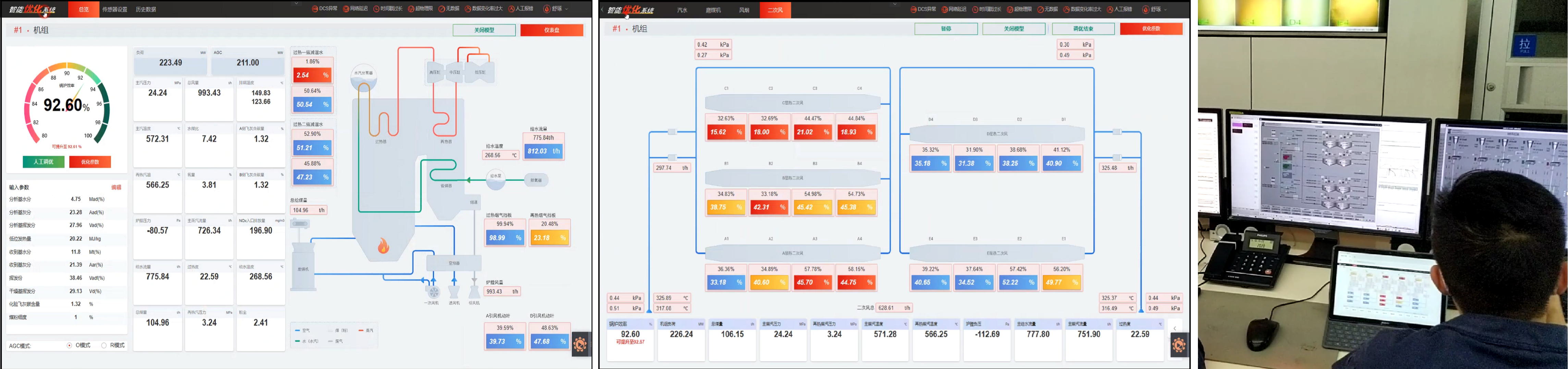}
	}
	\subfigure[System deployed in Shanxi Xingneng Power Station]{
		\includegraphics[width=0.9\textwidth]{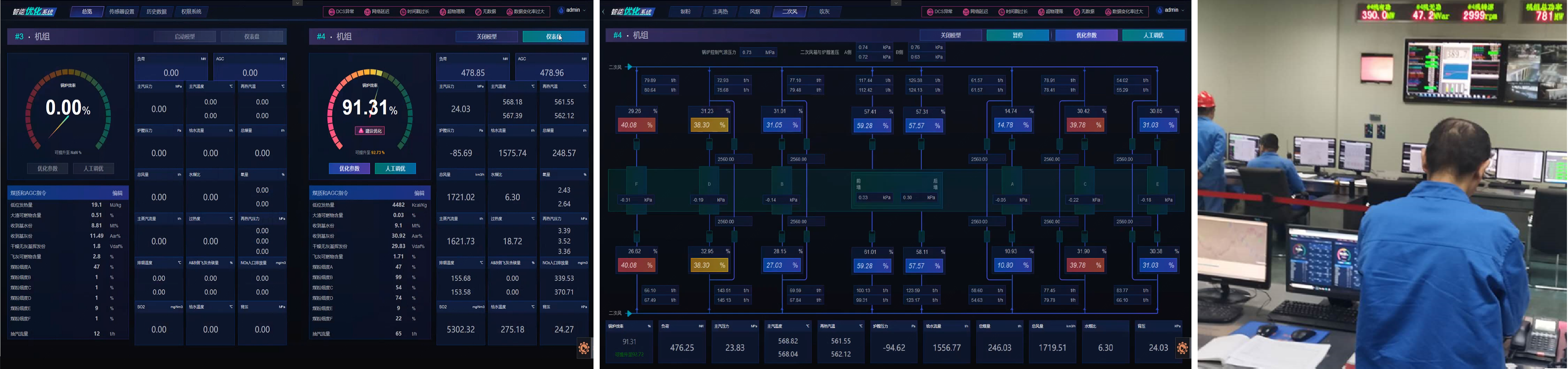}
	}
	\caption{System interfaces and control room usage of DeepThermal in two thermal power plants in China}
	\label{sys.inter}
\end{figure*}

\subsubsection{Detailed System Implementation}

In this section, we provide additional information for the DeepThermal systems deployed in real-world thermal power plants. 
Figure \ref{sys.inter} presents the DeepThermal systems deployed in CHN Energy Langfang Power Station and Shanxi Xingneng Power Station. 
The user interface of DeepThermal consists of an overview screen and several sub-interfaces displaying recommended control strategies for different combustion control stages (e.g. coal pulverizing, burning, air circulation and steamer system, etc.).
The leftmost figures in Figure \ref{sys.inter} show the overview screens of DeepThermal.
Information that are not obtainable from sensors can be manually inputted in the bottom-left part in the overview screen, such as the chemical property of the coal.
The figures in the middle are the sub-interfaces of the burning stage, which display recommended values for valves of the secondary blowers in the burner. Other sub-interfaces are not presented due to space limit. 
The rightmost figures in Figure \ref{sys.inter} show the scene that the operator using DeepThermal to adjust their control strategy in the central control room.

DeepThermal displays two lines of values for each control variable in the interface. The value in the top line is the current control value. The value in the bottom with red, yellow or blue color marks the recommended value from the learned RL policy. The red or yellow indicate that the current control strategy has a large or medium deviation from the optimal policy, which should be adjusted. The operator in the central control room can easily adjust his control operation following the guidance of this system. 

For each power plant, the user interface style of DeepThermal is customized to meet the needs of power plant clients. The interface layouts are also redesigned to match with the interface of the existing distributed control system (DCS) in the power plant. This allows operators easily locating the corresponding control element and adapting to the guidance from DeepThermal. Despite the differences in the system frontend, the RL algorithm module and system backend remain the same for different power plants.

\subsubsection{Extra Real-World Experiments}

\begin{figure*}[th]
	\centering
	\subfigure[250 MW experiment on 2020/07/17]{
		\label{fig:subfig:a}
		\includegraphics[width=0.66\columnwidth]{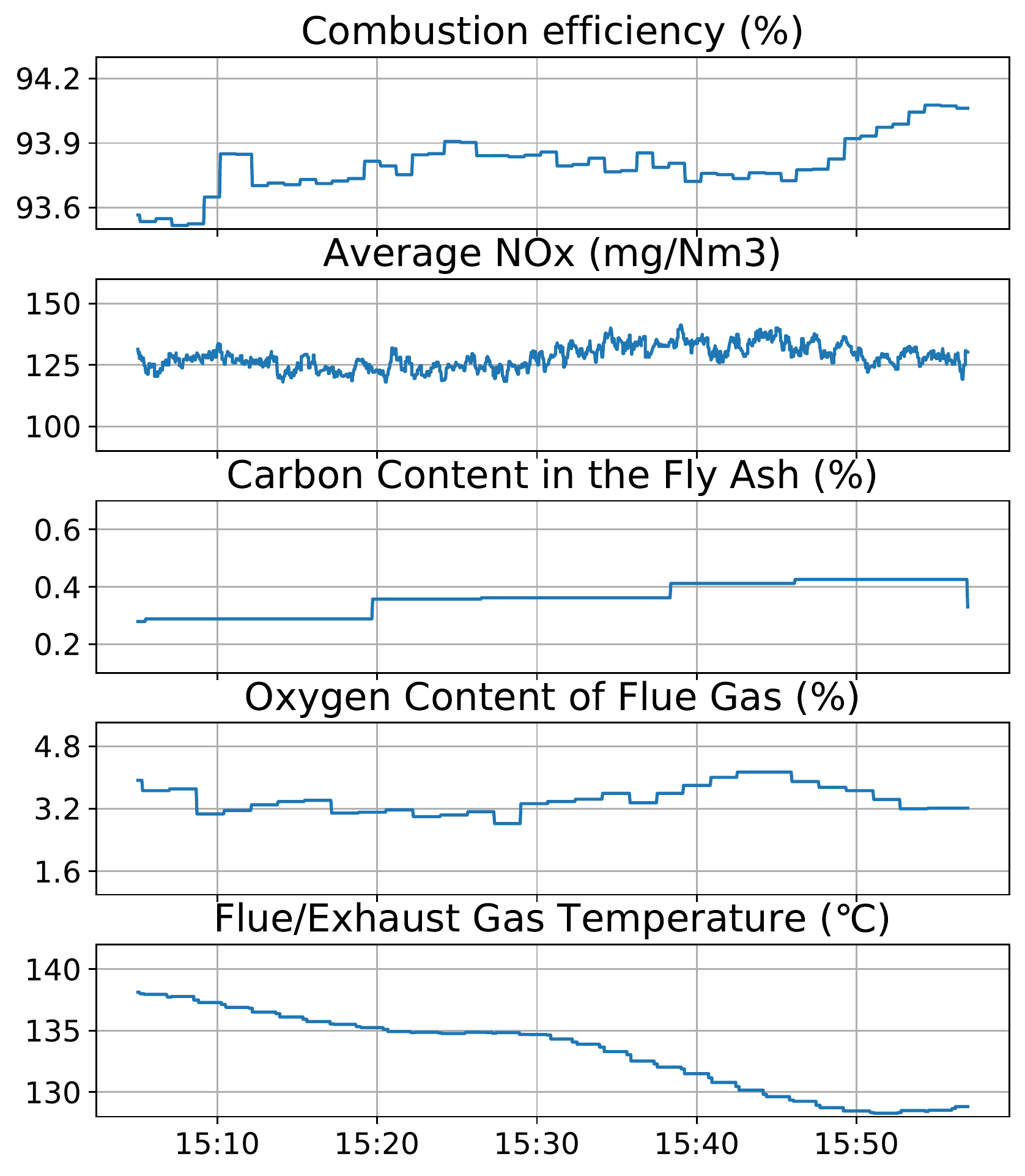}
	}
	\subfigure[200MW experiment on 2020/07/18]{
		\label{fig:subfig:b}
		\includegraphics[width=0.66\columnwidth]{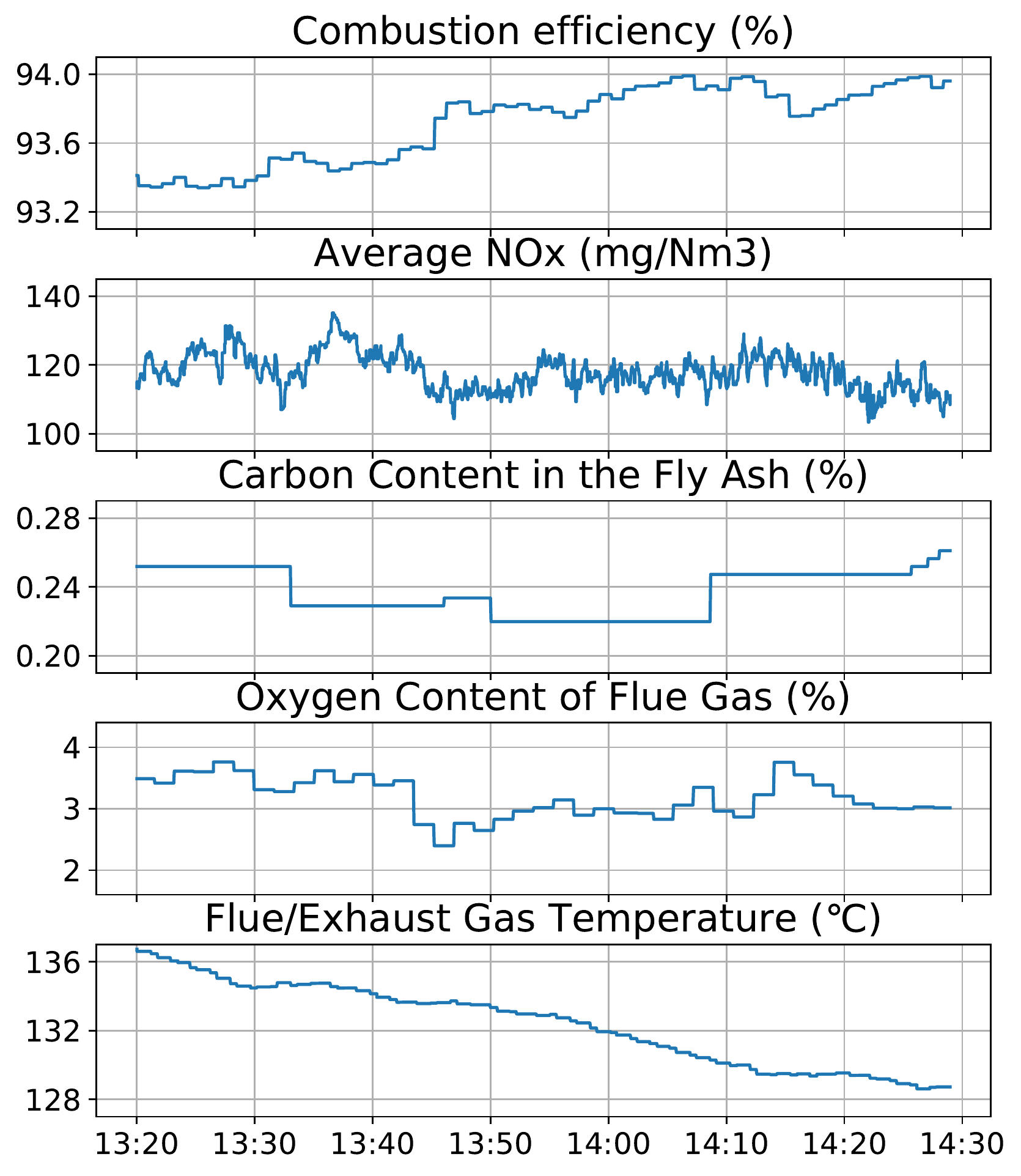}
	}
	\subfigure[300MW experiment on 2020/07/22]{
		\includegraphics[width=0.66\columnwidth]{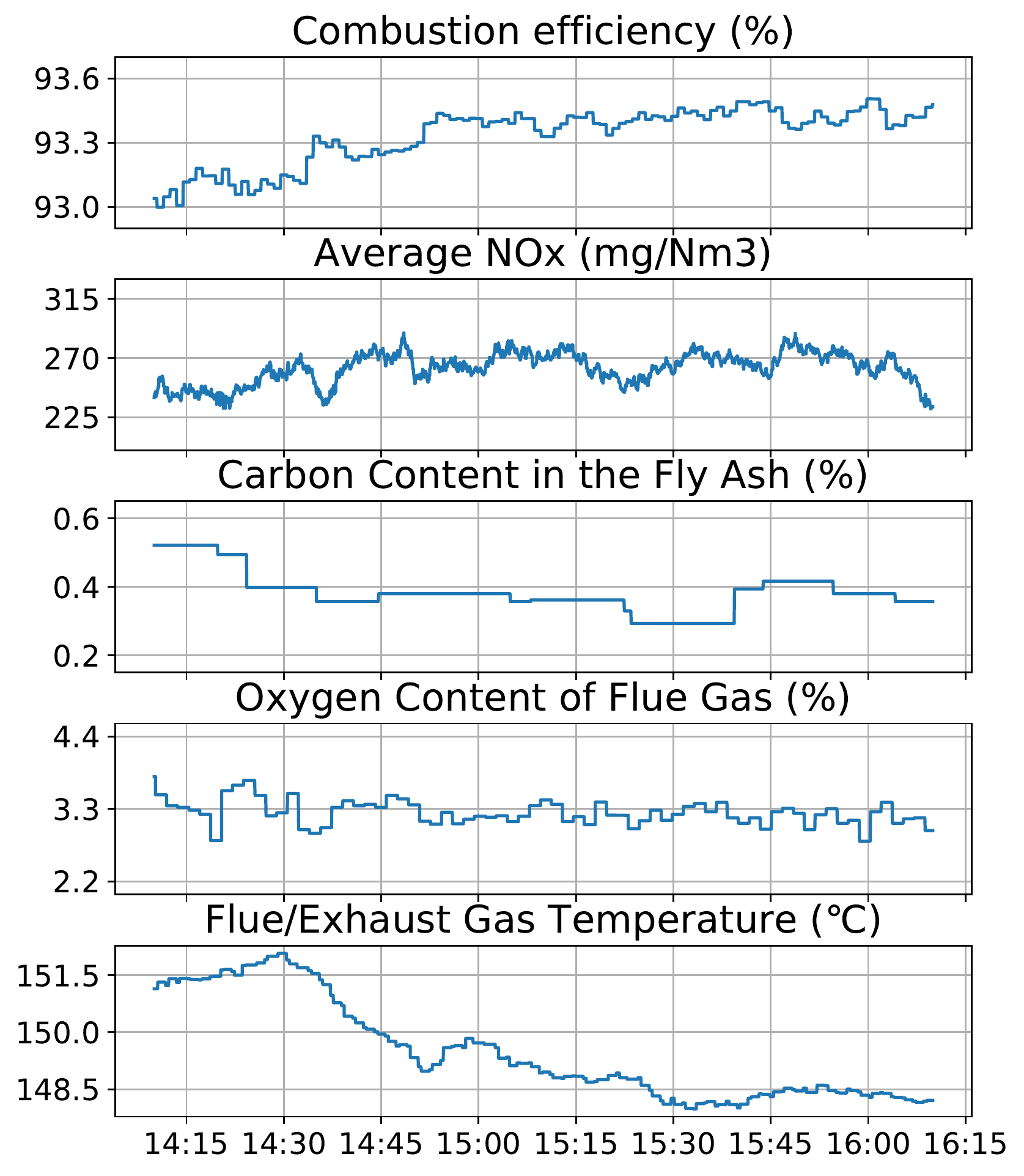}
	}
	\caption{Real-world experiments at CHN Energy Langfang Power Station}
	\label{fig:langfang} 
\end{figure*}

In this section, we report results from recent real-world experiments conducted at CHN Energy Langfang Power Station in Figure \ref{fig:langfang}. DeepThermal system uses more than 700 sensors in the TPGU and optimizes the control strategy involving more than 70 control variables.

Experiment (a) was conducted in the 250MW load setting on July 17, 2020. The test started at 15:20, and ended at 16:23. The optimized control strategy achieved the maximum increase of 0.56\% on the combustion efficiency and the maximum decrease of 9.8\textcelsius{} on the exhaust gas temperature in about 60 minutes compared with the initial values. The average $\mathrm{NO_x}$ concentrations, the oxygen content of flue gas remained at a relatively stable level. 

Experiment (b) was carried out at 14:36-15:55 on July 18, 2020 in the 200MW load setting. The initial value of combustion efficiency was 93.34\%., and it reached 93.99\%  after the adjustment in about 60 minutes, the optimized control strategy achieved a maximum increase of 0.65\%. The average $\mathrm{NO_x}$ concentrations reduced from $128.46 mg/Nm^3$ to $118.61 mg/Nm^3$, which achieved the maximum decrease of about $10 mg/Nm^3$. 
The exhaust gas temperature also dropped from 136.7\textcelsius{} to 128.1\textcelsius{}. Other indicators also showed a certain level of decrease.

Experiment (c) was conducted in the 300MW load setting on July 22, 2020. The test started at 14:10 and ended at 16:15. After optimization, the combustion efficiency rose from 93\% to 93.51\%, which achieved the maximum increase of about 0.51\%; The carbon content in the fly ash decreased from 0.56\% to 0.38\% in about 60 minutes.
In all the three different load settings (200MW, 250MW,300MW), DeepThermal effectively improved combustion efficiency and decreased the values of the other three key indicators.

\subsection{Algorithm Implementation Details}

\subsubsection{Real-sim data combination in MORE.}
\begin{figure}[t]
	\includegraphics[width=0.47\textwidth]{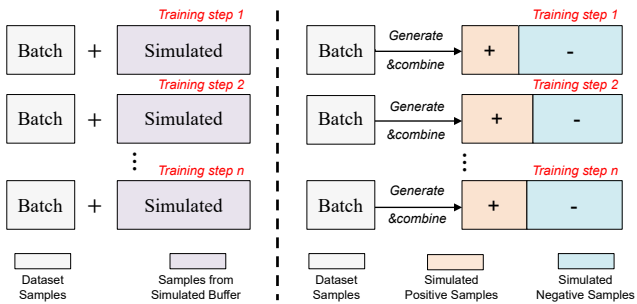}
	\centering
	\caption{Illustration of hybrid training in MORE}
	\label{fig_hyprid} 
\end{figure}

Rather than na\"ively mixing real and simulated data as shown in the left part of Figure \ref{fig_hyprid}, MORE constructs a special local buffer $\mathcal{R}$ to combine real, positive and negative simulated data for training (right figure). 
As DeepThermal uses a RNN-based simulator, to improve the prediction accuracy, we first use trajectories in the real data batch to preheat and update the hidden states of the simulator, and then use it to roll out trajectories.
We also control the proportion of positive and negative samples in the simulated data with the percentile threshold $\beta_p$. The choice of $\beta_p$ controls the behavior of MORE to be either more aggressive to explore beyond the offline dataset (with a small $\beta_p$) or more conservative to avoid OOD errors (with a large $\beta_p$).

\subsubsection{Hyper-parameters for Real-World Experiments.}
For the combustion process simulator, each LSTM cell has 256 hidden units. For the sequence to sequence structure, we set the encoder length as 20, and the decoder length as 10. The model is trained with the Adam optimizer and a learning rate of $1\mathrm{e}-4$. 
In the scheduled sampling, the probability of replacing the true data with the generated ones decays by $1\mathrm{e}-4$ for every training step. For the noisy data augmentation, the initial noises are sampled from $N(0, 0.025)$.
The actor, critics and the state-action VAE are modeled using fully connected neural networks, which are optimized using Adam with learning rates being $1\mathrm{e}-6$, $1\mathrm{e}-5$ and $1\mathrm{e}-3$ respectively. Both the encoder and the decoder of the state-action VAE have three hidden layers $(1024,1024,1024)$ by default. 
The policy and the critic networks have three hidden layers $(256,128,64)$. 
The state-action VAE, policy and critic networks are trained for $1$ million steps with batch size $256$.

\begin{table}[ht]
	\centering
	\footnotesize
	\caption{Rollout length $H$ used in D4RL benchmarks}
	\begin{tabular}{c|c|c|c|c|cc}
		\hline
		\textbf{Env}  & \multicolumn{2}{c|}{halfcheetah} & \multicolumn{2}{c|}{hopper} & \multicolumn{2}{c}{walker2d}        \\ \hline
		\textbf{Type} & medium          & mixed          & medium        & mixed       & \multicolumn{1}{c|}{medium} & mixed \\ \hline
		\textbf{H}    & 1               & 1              & 5             & 5           & \multicolumn{1}{c|}{5}      & 1     \\ \hline
	\end{tabular}
	\label{tab:rollout}
\end{table}

\subsubsection{Hyper-parameters for Offline RL Benchmarks.}
For all function approximators, we use fully connected neural networks with RELU activations. 
The pre-trained dynamics model has four hidden layers $(200,200,200,200)$ with the learning rate of $1\mathrm{e}-4$.
The pre-trained state-action VAE has two hidden layers $(750,750)$ with the learning rate of $1\mathrm{e}-4$.
The policy network is a 2-layer MLP with 300 hidden units on each layer, and we use tanh (Gaussian) on outputs. The reward and cost critic networks are 2-layer MLPs with 400 hidden units each layer. 
The learning rates is $1\mathrm{e}-5$ for the policy network and $1\mathrm{e}-3$ for the Q-networks.
We use the soft update for the target Q-functions with the rate of $0.005$ per iteration.
We use Adam as the optimizer for all networks. The batch size is 256 and $\gamma$ is 0.99.
We search the model sensitivity threshold $\beta_u \in \{40, 70\}$ and the data density threshold $\beta_p \in \{10, 40, 70\}$. The rollout length are given in Table \ref{tab:rollout}. We search the rollout length $H \in \{1, 5\}$ and use the penalty coefficient of $\kappa=5$ in the experiments.

\end{document}